# Hybrid supervised instance segmentation by learning label noise suppression

Linwei Chen [a], Ying Fu [a,*], Shaodi You [b], Hongzhe Liu [c]

[a] School of Computer Science and Technology, Beijing Institute of Technology, Beijing 100081, China
[b] Computer Vision Research Group, Insititute of Informatics, University of Amsterdam, Amsterdam 1098 XH, Netherlands
[c] Beijing Key Laboratory of Information Service Engineering, Beijing Union University, Beijing 100101, China

ABSTRACT

To reach top accuracy, current fully supervised instance segmentation methods severely rely on large-scale pixel-wise labeled datasets. They are usually expensive and time-consuming to obtain. Though weakly or semi-supervised methods utilize cheap bounding box labeled, image-level labeled or unlabeled samples to save the labeling cost, their performance is largely sacrificed. To save labeling cost without losing much performance, in this paper, we present a pipeline that can utilize economical bounding box labels and accurate pixel-wise labels in a hybrid way. Specifically, we design two ancillary models to learn label noise suppression and obtain accurate pseudo pixel-wise labels from bounding box labels for training. One is designed to suppress mislabeling between foreground and background, and the other is designed to suppress noise from mislabeling of instances. Moreover, we exploit category-aware spatial attention module, category constraint module, instance constraint module, and self-learning training approach to improve the accuracy of pseudo labels. Experiments on the PASCAL VOC 2012 and the Cityscapes datasets show that our method can achieve competitive performance with much less labeling cost.

## 1. Introduction

Instance segmentation is a fundamental task in computer vision and artificial intelligence. It plays a crucial role in autonomous driving [16], robotics [19], medical image diagnosis [55] and *etc.*

Recently, fully supervised instance segmentation methods [23,24,45,11,38] have been intensively studied and they profit from a large amount of training data to reach top performance. However, their accuracy heavily relies on the quantity and quality of pixel-wise labeled training data, which is labor-intensive and expensive to create [18,42,14]. To be precise, drawing a segmentation on an object needs 79 s [6] while obtaining a bounding box needs only 7 s [48], which is about 11.3× more expensive.

To reduce labeling cost, some weakly supervised methods [64,21,1,66,13,30,39,26,4] and semi-supervised methods [7] have been proposed. Instead of using pixel-wise labeled samples, they utilize economical image-level labeled [64,21,1,66,13], bounding box labeled [30,39,26,4] or unlabeled samples [7] for learning. Among them, weakly supervised methods [26,4] using bounding box supervision achieve the best performance. Nevertheless, they still have a large performance gap compared with fully supervised methods [24,38], which is unacceptable in practical usage. We attribute this to strong and notorious label noise in bounding box labels, and existing works [26,4] cannot handle it well. Specifically, we identify two types of noise in bounding box labels for instance segmentation, *i.e.*, background noise and instance noise, as illustrated in Fig. 1. Background noise mainly comes from the uncertain foreground and background separation within each bounding box. Instance noise is caused by the overlap of multiple bounding boxes, resulting in confusion when assigning pixels in the overlapping area to different instances. Therefore, suppressing these two types of label noise in bounding box labels remains the key problem for achieving decent performance with low labeling cost.

To solve this problem, in this paper, we propose a novel hybrid supervised pipeline that combines full supervision and weak supervision. As illustrated in Fig. 2, the proposed pipeline consists of three models, including a semantic ancillary model, an instance ancillary model, and a primary instance segmentation model. The semantic ancillary model is designed to suppress background noise by learning semantic segmentation. And the instance ancillary model is designed to further suppress instance noise by learning instance segmentation. They can be easily optimized with a small subset of clean pixel-wise labels (*e.g.*, 1%) in a fully supervised way, which only increases very little labeling cost and ensures the accuracy of label noise suppression. Then, any instance segmentation model can be trained with bounding box labels in a weakly supervised way. Though the whole idea is straightforward, there are two challenging problems for us to achieve decent performance. First, how to fully utilize rich information in bounding box labels to correct pseudo labels? Second, how to avoid overfitting when the number of pixel-wise labels is limited?

* Corresponding author.
*E-mail address:* fuying@bit.edu.cn (Y. Fu).

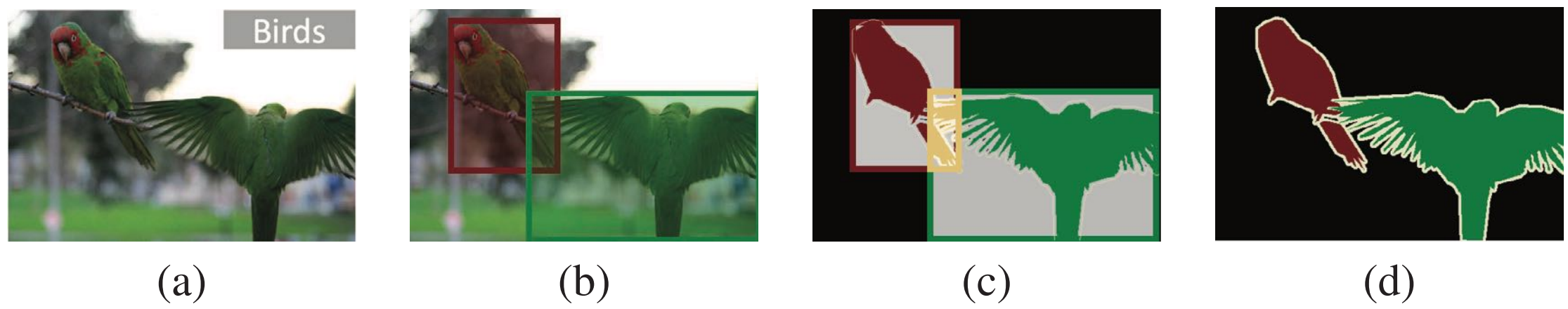

**Fig. 1.** (a) Image-level label lacks instance-level information, and locations of objects are uncertain. (b) Shapes of objects are uncertain in bounding box labels. (c) Two types of noise in bounding box labels, background noise is marked as gray, and instance noise is marked as yellow in the overlapping area. (d) Instance-level pixel-wise label (Ground truth for instance segmentation).

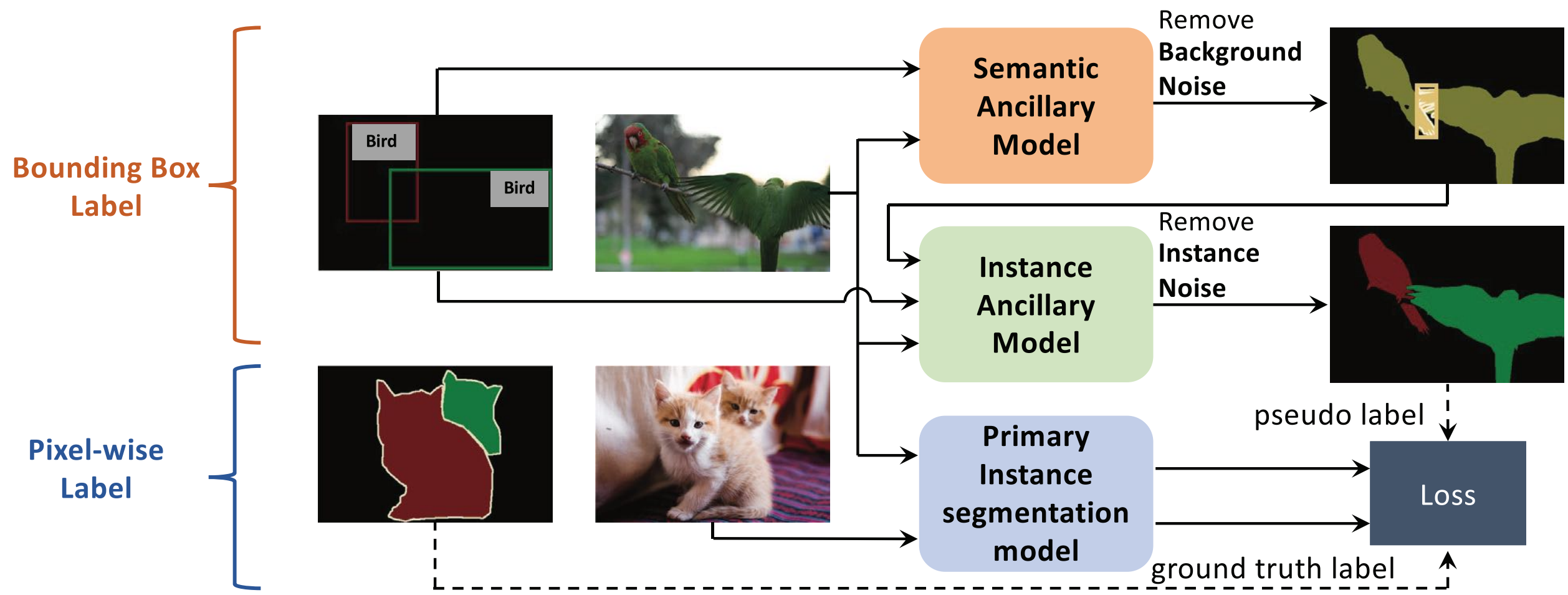

**Fig. 2.** Overview of the proposed method. The semantic ancillary model in the first step is designed to suppress the background noise in bounding boxes. The instance ancillary model in the second step is designed to suppress the instance noise. Then, the produced pseudo instance-level pixel-wise labels can be used as ground truth to train the primary instance segmentation model in the third step. Notice that the pipeline is only involved in the training stage, and the primary instance segmentation model works independently after training.

To address the first challenge, we propose a category-aware spatial attention module, a category constraint module, and an instance constraint module. The category-aware spatial attention module directly fuses information of bounding box labels and intermediate semantic segmentation results (*i.e.*, information of category, location, and shape of objects) into the deep features of CNN, helping ancillary models to generate more accurate pseudo labels. The category constraint module and instance constraint module utilize the label consistency prior (*i.e.*, the segmentation mask should be inside the bounding box) to refine the segmentation results. For the second challenge, we exploit the self-learning training approach. During training, it utilizes pseudo labels generated from bounding box labeled images by ancillary models to optimize themselves. To avoid introducing noise, we ignore loss from the noisy area, *e.g.*, an area where two bounding boxes overlap. The self-learning training approach can not only avoid overfitting with limited pixel-wise labels but also improve the label noise suppression performance.

Experiments on the PASCAL VOC 2012 [18] and Citycapes [14] show that our hybrid supervised method can largely boost the performance compared with corresponding fully supervised baseline, and outperform some former state-of-the-art fully supervised instance segmentation methods, even with only 1% pixel-wise labels. We also extend our method to semantic segmentation task on the PASCAL VOC 2012 [18] and achieve better performance than a fully supervised setting with more than 75% reduction to the labeling cost.

The main contributions of this work can be summarized as follow:

- We present a pipeline that combines the advantages of the low labeling cost from bounding box labels and high accuracy from pixel-wise labels. It can fully utilize both types of labels in a hybrid way and well balance the performance and labeling cost.
- We design a semantic ancillary model and an instance ancillary model, which can effectively suppress background noise and instance noise in bounding box labels.
- We exploit category-aware spatial attention module, category constraint module, instance constraint module, and self-learning training approach in ancillary models to fully utilize bounding box labels and avoid overfitting with limited pixel-wise labels.

## 2. Related Work

In this section, we first review semantic segmentation methods and instance segmentation methods, then we introduce semi-supervised and weakly supervised instance segmentation methods.

**Semantic segmentation.** As a dense-prediction task, semantic segmentation assigns each pixel in an image to a specific category.

Fully convolutional networks (FCNs) [44] are the milestone in semantic segmentation. Later, a series of fully supervised methods [58,37] improve the accuracy of semantic segmentation by using contextual information [20,28,65], exploiting the multi-scale information [12,61,56,62], enhancing the fusion of features from different levels [60,59], or learning boundary information of targets [53]. Our ancillary models are based on one of these works, *i.e.*, DeepLabv3+ [12], which is reliable and easy to train.

**Instance segmentation.** Mainstream instance segmentation methods [24,11,9,10,38] first detect the objects with bounding boxes and then describe their shapes in binary masks. Another branch of instance segmentation methods [45,35,46,47] rely on dense-prediction, which first assigns a semantic category to each pixel, and then separates pixels into different instances. Currently, the state-of-the-art results are usually achieved by two-stage methods [24,11], which are based on reliable two-stage detectors like Faster R-CNN [49]. With the development of one-stage detectors [41,54], instance segmentation methods based on one-stage detector can achieve competitive results both in speed and performance [9,38,10]. To verify the effectiveness of our method, we choose a two-stage based method (*i.e.*, Mask R-CNN [24]) and a one-stage based method (*i.e.*, CenterMask [38]) as the primary instance segmentation model in our pipeline.

**Semi-supervised instance segmentation.** Though fully supervised instance segmentation methods [24,11,9,10,38,45] achieve reasonable performance, the labeling cost is unbearable. To address this problem, Bellver et al. [7] present a semi-supervised method for instance segmentation. It predicts pseudo labels on unlabeled images to augment the dataset. However, the pseudo-labels are often inaccurate for outlier examples, the noise and false positives in pseudo labels limit their performance. Li et al. [39] utilize both pixel-wise labels and bounding box labels for semi-supervised instance segmentation. It relies on [2,50] to obtain pseudo pixel-wise labels from bounding box labeled images, which impedes them to achieve similar performance with full supervision. A series of recent researches [27,36,63] aim to predict object masks for categories whose shapes are unknown in the dataset. They assume costly pixel-wise labels are available for a subset of categories and other categories have only bounding box labels. Although they make good progress in this task, the significant difference in shape between different categories limits their performance.

**Weakly supervised instance segmentation.** To save labeling cost, a few weakly supervised methods have been proposed, and they usually use image-level labels [64,21,1,66,13] or bounding boxes [30,26,39,4]. For image-level supervision, Zhou et al. [64] make use of class peak responses to locate objects and extract instance masks with some off-the-shelf methods [2]. Ge et al. [21] refine image-level labels to pixel-wise labels with a cascaded pipeline that is trained alternatively with a curriculum learning strategy. Ahn et al. [1] exploit the inter-pixel relations on class attention maps for training and discover instances by estimating rough locations of objects and detecting boundaries between different instances. Zhu et al. [66] use a novel instance extent filling approach to learn the full extent of pixel-level instance response. Cholakkal et al. [13] construct an object category density map to provide both the global object count and the spatial distribution of object instances. Though they largely reduce the labeling cost by using image-level labels and achieve promising results, their performance in instance segmentation is still much lower than fully supervised methods.

As for bounding box supervision, Khoreva et al. [30] and Li et al. [39] utilize GrabCut [50] and MCG [2] to propose instance masks from bounding box for learning. Hsu et al. [26] train the segmentation model with bounding box labels in an end-to-end way with multiple instance learning strategy. Though their results are better than the image-level supervised method, they can only partially suppress the background noise and cannot deal with the instance noise. These limit their performance. To balance the performance and labeling cost, we utilize bounding boxes for saving the labeling cost while maintaining a small portion of pixel-wise labels for learning label noise suppression.

## 3. Methodology

In this section, we first introduce our motivation and formulation, then we describe our pipeline and details of the proposed category-aware spatial attention module, category constraint module, instance constraint module, and self-learning training approach. Finally, we introduce the loss function for learning.

### 3.1. Motivation and Formulation

Instance segmentation is a task that aims to jointly detect and separate each distinct foreground object of interest from the background. Our goal is to save the labeling cost while keeping the performance for instance segmentation. As shown in Table 1, both image-level label and bounding box label can largely save the labeling cost, but image-level label lacks instance-level information which is crucial for achieving satisfactory performance, therefore bounding box label is a better choice for our goal.

Obviously, compared with ground truth instance-level pixel-wise label $Y$, the bounding box label $B$ is not accurate enough and contains two types of label noise, *i.e.*, background noise and instance noise as shown in Fig. 1, which inevitably leads to inferior performance of weakly supervised methods [30,39,26,4]. Taking an image of two object instances as an example, let $0, 1$ and $2$ represent background, , foreground instance 1 and instance 2, respectively. In the ground truth instance-level pixel-wise label $Y$, each pixel $x$ in image $X$ is assigned to the correct category and instance. But for bounding box label $B$, some background pixels may be labeled with foreground category, and pixels belonging to one of the instances may be assigned to the other incorrectly, *i.e.*,

$$\begin{aligned} &P(b>0|y=0)=\rho_0, \quad 0\leqslant\rho_0\leqslant 1 \\ &P(b=1|y=2)+P(b=2|y=1)=\rho_1, \quad 0\leqslant\rho_1\leqslant 1 \end{aligned} \tag{1}$$

where $b, y \in \{0,1,2\}$ are labels for each pixel in $B$ and $Y$. $\rho_0$ indicates the proportion of background pixels that are assigned to the foreground category, and shows the background noise level. $\rho_1$ indicates a proportion of foreground pixels that are assigned to incorrect instances and shows the instance noise level.This analysis also applies to images with more categories and instances.

Previous works [30,39,26,4] usually rely on hand-crafted heuristics [2,50] to suppress label noise and generate pseudo label with lower $\rho_0$ and $\rho_1$, but generated pseudo label is not accurate enough as shown in Fig. 3 (b), (c) and Table 2. To solve this, we propose a semantic ancillary model and instance ancillary model, they can learn to suppress background noise and instance noise by utilizing a small subset of clear pixel-wise labeled samples. There have been some existing works [39,52,29] experiment with pixel-wise labels and bounding box labels. However, [52,29] are designed only for semantic segmentation task and cannot deal with instance segmentation task, while [39] still rely on hand-crafted heuristics [50,2] for training and cannot make use of pixel-wise labeled samples to refine the pseudo labels, which limits its accuracy. As illustrated in Fig. 2, the proposed method is able to utilize pixel-wise labels and bounding box labels in a hybrid way. Figs. 3 (d), (e) and Table 2 show the label noise in generated pseudo pixel-wise labels are largely suppressed, and it is visually very close to ground truth.

**Table 1**
The labeling cost (in seconds) estimation of heterogeneous labels on the PASCAL VOC 2012 [18] from previous works [6,7], and it can also be applied to the Cityscapes [14].

| Label type | PASCAL VOC 2012 [18] | Cityscapes [14] |
|---|---|---|
| Image-level | 20.0s/image | 8.0s/image |
| Bounding box | 38.1s/image | 127.5s/image |
| Pixel-wise | 239.7s/image | 1,387.5s/image |

### 3.2. Pipeline

As illustrated in Fig. 2, we organize our pipeline in a three-step cascade way. In the first step, we aim to suppress the background noise in bounding box labels, *i.e.*, separate foreground pixels from the background in the bounding box. We view it as a semantic segmentation task and design a semantic ancillary model based on DeepLabv3+ [12]. In this step, our semantic ancillary model can produce accurate semantic segmentation result $Y'$:

$$Y' = f_{\theta}(X, h_{\psi}^{CASA}(B)), \tag{2}$$

where $f, h^{CASA}$ are the semantic ancillary model and category-aware spatial attention module, and $\boldsymbol{\theta}, \boldsymbol{\psi}$ are their corresponding parameters.

In the second step, we aim to suppress the instance noise in bounding box labels, *i.e.*, assigning foreground pixels to correct instances. We view it as an instance segmentation task and design an instance ancillary model based on DeepLabv3+ [12] to predict pseudo labels in high resolution. But the spatial-invariant property of fully convolutional networks makes it unable to identify different instances in different spatial locations. To solve this, we change the decoder of DeepLabv3+ [12] into two branches and utilize spatial-embedding loss introduced by Neven et al. [45] for training. The spatial-embedding loss [45] assigns each pixel a spatial coordinate to avoid the spatial-invariant problem, and helps the model to learn to identify different instances by predicting position-relative offset vectors that point to its corresponding instance center. Furthermore, the instance segmentation results are non-overlapping, so the instance noise introduced by overlapping instance proposals is avoided.

As shown in Fig. 4 (b), the instance ancillary model has two decoder branches, *i.e.*, confidence branch and instance branch. They mainly consist of two deconvolutional layers, each decoder is for $2\times$ upsampling, which is a total $4\times$ upsampling scale. Two deconvolutional layers are connected by a residual block [25]. The first deconvolutional layer is followed by a batch normalization layer and a ReLU layer. The second deconvolutional layer is followed by a sigmoid layer or tanh layer for predicting confidence maps or offset maps respectively.

The instance branch predicts pixel offset maps and margin maps. Pixel offset maps contain pixel offset values $(o_{im}, o_{in})$ for obtaining the predicted instance center $(c_{im}, c_{in})$:

$$\begin{aligned} c_{im} &= o_{im} + e_{im}, \\ c_{in} &= o_{in} + e_{in}, \end{aligned} \tag{3}$$

where $(e_{im}, e_{in})$ are the coordinates of pixels. The predicted centers of pixels should be close to each other if they belong to the same instance, which can be recognized by spatial clustering. Considering sizes and shapes of instances are different, if clustering margin is kept the same for all instances, closely distributed small instances may be clustered into one object because the margin is relatively large for them, or large instance may be clustered into multiple instances because the pixels far from the center may be unable to point into this relatively small margin region around the instance center. To avoid these bad cases, the instance branch predicts margin maps that contain instance-specific clustering margin values $(\sigma_{im}, \sigma_{in})$ in width and height direction, which are adapted to the shape and size of instances during learning.

The confidence branch is used to obtain category-specific confidence maps. The instance ancillary model is trained to predict higher confidence value $d_i$ for a foreground pixel $i$ if its prediction of $(c_{im}, c_{in})$ and $(\sigma_{im}, \sigma_{in})$ are closer to the ground truth center $(\bar{C}_{km}, \bar{C}_{kn})$ and margin values $(\bar{\sigma}_{km}, \bar{\sigma}_{kn})$ of a category-specific instance $S_k$. And a pixel with confidence value $d_i \leqslant 0.5$ is regarded as a background pixel.

During inference, for each category-specific bounding box, the foreground pixel $i'$ with highest confidence value $d_{i'}$ is chosen, its predication of $(c_{i'm}, c_{i'n})$ and $(\sigma_{i'm}, \sigma_{i'n})$ are regarded as estimated center $(\hat{C}_{km}, \hat{C}_{kn})$ and margin values $(\hat{\sigma}_{km}, \hat{\sigma}_{kn})$ of instance $S_k$ respectively. By using this center and accompanying margin values, we cluster foreground pixel $i$ into instance $S_k$ if the predicted center $(c_{im}, c_{in})$ of pixel $i$ is close to $S_k$. The distance is defined by the gaussian function, *i.e.*,

$$i \in S_k \iff \exp\left(-\frac{\left(c_{im} - \hat{C}_{km}\right)^2}{2\hat{\sigma}_{km}^2} - \frac{\left(c_{in} - \hat{C}_{kn}\right)^2}{2\hat{\sigma}_{kn}^2}\right) > 0.5. \tag{4}$$

In this step, the instance ancillary model can suppress instance noise in semantic segmentation result from the first step and produces accurate pseudo instance-level pixel-wise instance labels $Y''$:

$$Y'' = g_{\phi}(X, h_{\omega}^{CASA}(B, Y')), \tag{5}$$

where $g, h^{CASA}$ are instance ancillary model and category-aware spatial attention module, and $\boldsymbol{\phi}, \boldsymbol{\omega}$ indicate their parameters. In $Y''$, shapes of objects within bounding boxes are largely recovered, *i.e.*, most of pixels that have incorrect labels in $B$ are assigned to correct categories and instances.

In the third step, we employ vanilla Mask R-CNN [24] and CenterMask [38] as the primary instance segmentation model. They are trained with both ground truth pixel-wise labels and pseudo pixel-wise labels obtained from bounding box labels.

### 3.3. Category-Aware Spatial Attention Module

Bounding box labels contain rich information about the existence of each category and the location of objects, which can be a great help for generating accurate pseudo pixel-wise labels. Existing box-level weakly supervised methods usually use them

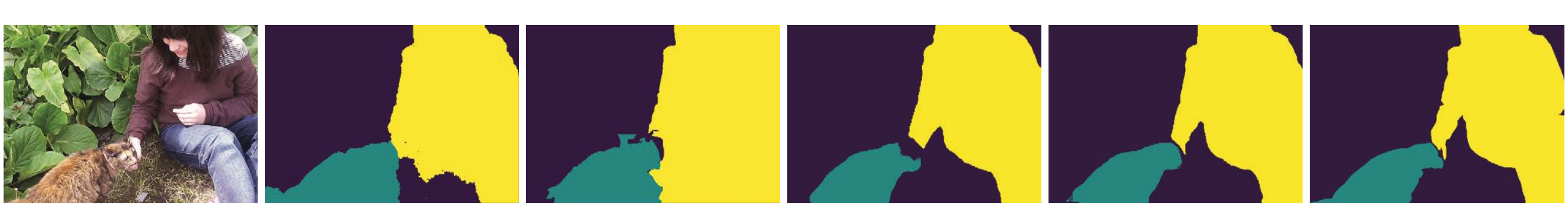

(a) Image (b) Grabcut (c) MCG (d) Ours (1%) (e) Ours (10%) (f) Ground truth

**Fig. 3.** Examples of pseudo labels on a specific image of the PASCAL VOC 2012 [18]. Different colors indicate differe.nt instances.

**Table 2**
Pseudo pixel-wise label quality evalutation on the PASCAL VOC 2012 [18] validation set. We follow [30] to use $mAP^r_{0.5}$ and $mAP^r_{0.75}$ to evaluate the accuracy of pseudo labels and level of label noise. Our method is trained with 1% (105 samples) and 10% (1058 samples) pixel-wise labels.

| Method | $mAP^r_{0.5}$ | $mAP^r_{0.75}$ |
|---|---|---|
| Rectangle box | 21.6 | 1.8 |
| Ellipse | 29.5 | 3.9 |
| MCG [2] | 28.3 | 5.9 |
| GrabCut [50] | 38.5 | 13.9 |
| Ours (1%) | 64.2 | 39.1 |
| Ours (10%) | 70.2 | 49.3 |

to refine inaccurate segmentation proposals [30,39,4] or utilize them as noisy supervisory signals [26], which is not optimal.

To fully utilize the information in bounding box labels and intermediate semantic segmentation results, we propose a novel category-aware spatial attention (CASA) module to fuse bounding box labels into the deep feature of ancillary models. It consists of two CASA blocks, as shown in Fig. 4 (c), they compute attention maps for 4× and 16× downsampling scales respectively to better adapt to objects of different sizes. As shown in Fig. 4 (c), there are two branches in the CASA block, the upper is the category branch and the bottom is the shape branch.The category branch is designed to explicitly obtain category information in bounding box labels and semantic segmentation results by using global pooling (*i.e.*, global max pooling and global average pooling) and a multi-layer perceptron (MLP). It produces channel attention tensor $\mathscr{F}^c \in \mathbb{R}^{N\times 1\times 1}$ ($N$ is the channel number of backbone feature map), which can adaptively choose and suppress channel features of backbone based on the category information. The shape branch is designed to obtain shape information in bounding box labels and semantic segmentation results. The $1\times 1$ convolutional layer and sigmoid layer can produce point-wise spatial attention tensor $\mathscr{F}^s \in \mathbb{R}^{N\times H\times W}$ ($N, H, W$ indicate channel, height, width dimensions of backbone feature map). It is well known that object boundaries are usually hard for instance ancillary model to estimate accurately, and they easily introduce label noise. To make instance ancillary model pay more attention to object boundaries, we add an extra $3\times 3$ convolutional layer to explicitly obtain the edges information from semantic segmentation results. In this way, the CASA module can fully extract the information of both bounding box labels and segmentation results, and fuse them into the corresponding feature map of ancillary models using element-wise multiplication. As shown in Fig. 5 (a) and (b), the CASA module enhances the feature of foreground objects and helps ancillary models to largely avoid incomplete segmentation.

The input of CASA blocks in the semantic ancillary model is a 3-dimensional tensor representing the bounding box label called label maps, where if a pixel belongs to a bounding box with a specific category, its corresponding category channel is set to 1 and the background channel is set to 0. Notice that a pixel can belong to multiple bounding boxes. As for instance ancillary model, the intermediate segmentation results are also concatenated to the input of CASA blocks.

### 3.4. Category Constraint Module & Instance Constraint Module

Considering generated pseudo pixel-wise label should be consistent with the bounding box label, *i.e.*, the segmentation mask should be inside the bounding box, we propose a category constraint module and instance constraint module to further improve the accuracy of pseudo labels.

As shown in Fig. 4 (c), the category constraint module can constrain category solution space for each pixel, *e.g.*, a pixel inside a bounding box without overlapping could only belong to the background or category of the bounding box. It avoids misclassification inside the bounding box and largely removes false positives outside the bounding box as illustrated in Fig. 5 (c).

Meanwhile, the original clustering procedure [45] of instance ancillary model may lead to inaccurate segmentation results when objects are too close to each other, *i.e.*, two objects can be clustered as one. To solve this, as shown in Fig. 4 (d), instance constraint module can filter out the area of confidence map outside the bounding box for each object sequentially before the clustering procedure, so as to reduce instance noise. Fig. 5 (d) shows it helps the ancillary model to better generate pseudo labels for different instances.

### 3.5. Self-Learning

Samples with clear pixel-wise labels are important for ancillary models to learn label noise suppression, and the number of samples should be small to keep total labeling cost low. However, insufficient training samples may result in the overfitting problem. To solve this, we utilize pseudo labels generated from bounding box labeled images by ancillary models to optimize themselves during training. It works similar to iterative label refinement mechanism [30,39]. To avoid introducing noise from pseudo labels, we adopt a strategy of ignoring noisy area, *i.e.*, we ignore the losses from the pixels of uncertain categories (whose max confidence is less than 0.5 in pseudo labels) and pixels of uncertain instances (whose locations are in overlapping bounding box area). As shown in Fig. 5 (e) and (f), the self-learning training approach not only solves the overfitting problem but also improves the quality of pseudo labels.

### 3.6. Loss Function

During training, the semantic ancillary model can be easily optimized with cross entropy loss. As for the instance ancillary model, confidence branch and instance branch are optimized jointly, the total loss is

$$\mathscr{L}_{total} = \lambda_{conf}\mathscr{L}_{conf} + \lambda_{inst}\mathscr{L}_{inst} + \lambda_{smooth}\mathscr{L}_{smooth}. \tag{6}$$

where $\lambda_{conf} = 1, \lambda_{inst} = 10, \lambda_{smooth} = 1$ for balancing losses.

The loss for confidence branch $\mathscr{L}_{conf}$ is a pixel-wise L2 loss

$$\mathscr{L}_{conf} = \frac{1}{H\times W}\sum_{i}^{H\times W}\mathbb{1}_{\{i\in S_k\}}\|d_i - \varphi_k(c_{im}, c_{in})\|^2 + \mathbb{1}_{\{i\in bg\}}\|d_i - 0\|^2, \tag{7}$$

where $H\times W$ is the number of pixels in image, $\mathbb{1}$ is indicator function, $d_i$ is confidence value, $bg$ represents background, and $\varphi_k(c_{im}, c_{in})$ is a two-dimensional gaussian function representing the distance to instance $S_k$ with probability

$$\varphi_k(c_{im}, c_{in}) = \exp\left(-\frac{(c_{im} - \bar{C}_{km})^2}{2\bar{\sigma}^2_{km}} - \frac{(c_{in} - \bar{C}_{kn})^2}{2\bar{\sigma}^2_{kn}}\right), \tag{8}$$

where $S_k$ is ground truth pixel-wise label if the sample is from pixel-wise labeled subset. And if the sample is from bounding box labeled subset, we regard all foreground pixels (whose predicted confidence value $d_i > 0.5$) inside each bounding box as pseudo label of $S_k$ for training *i.e.*, self-learning. Notice that we ignore losses from pixels of uncertain instances (whose locations are in overlapping bounding box areas) to avoid introducing instance noise. And $(\bar{C}_{km}, \bar{C}_{kn})$ is the center of instance $S_k$

$$\bar{C}_{km} = \frac{1}{|S_k|}\sum_{i\in S_k} c_{im}, \quad \bar{C}_{kn} = \frac{1}{|S_k|}\sum_{i\in S_k} c_{in}, \tag{9}$$

and $(\bar{\sigma}_{km}, \bar{\sigma}_{kn})$ is defined as

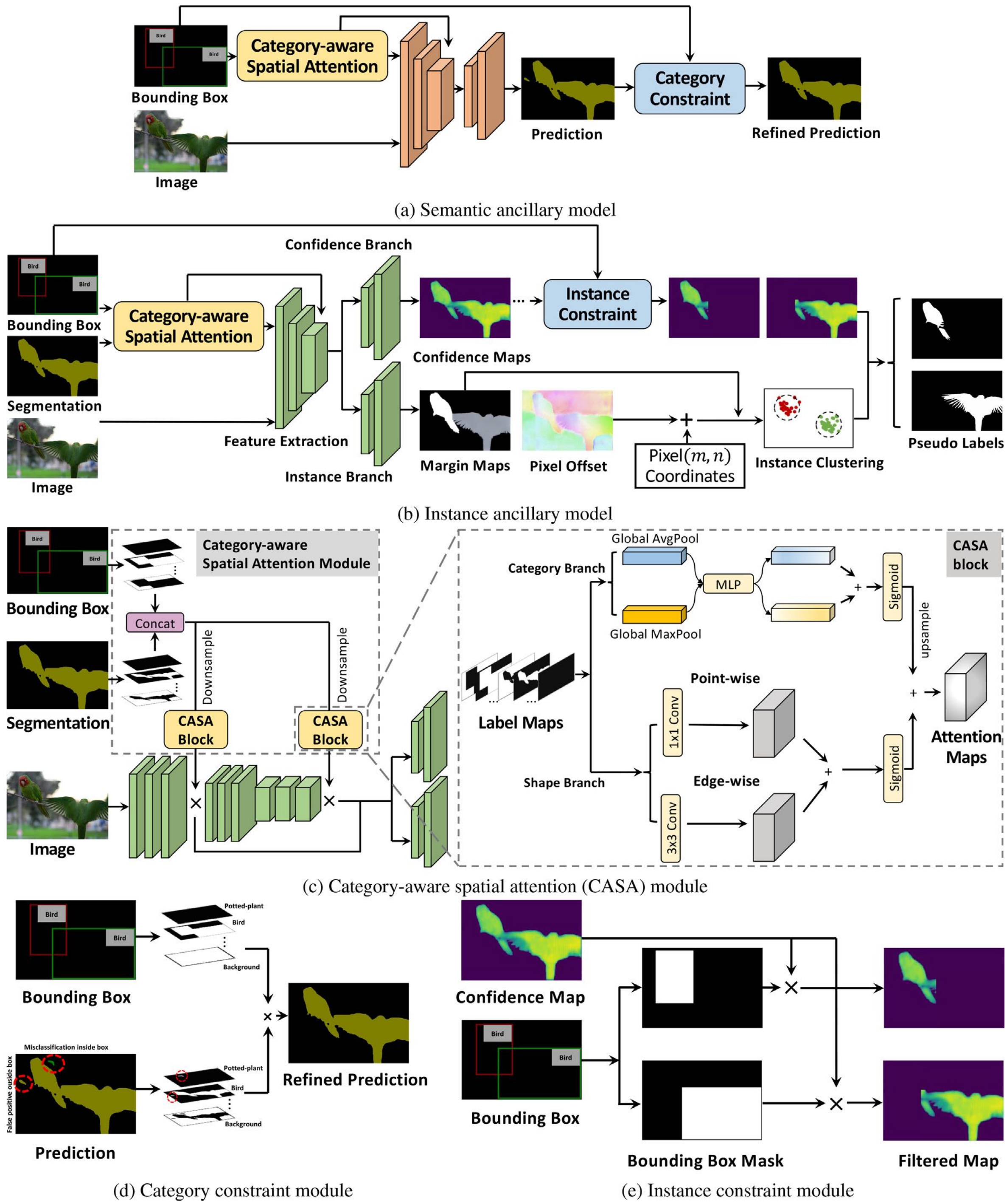

**Fig. 4.** (a) and (b) illustrate the semantic ancillary model and the instance ancillary model in our pipeline. In the pixel offset map of (b), the color indicates the angle of each pixel offset vector, and the saturation indicates the distance to the center. (c) describes details of category-aware spatial attention module in instance ancillary model, and (d) and (e) show how category constraint module and instance constraint module work in semantic ancillary model and instance ancillary model, respectively. The "+" in (b), (c) indicates pixel-wise addition, and the "×" in (c), (d), (e) indicates element-wise multiplication.

$$\bar{\sigma}_{km} = \frac{1}{|S_k|}\sum_{i\in S_k}\sigma_{im}, \quad \bar{\sigma}_{kn} = \frac{1}{|S_k|}\sum_{i\in S_k}\sigma_{in}. \tag{10}$$

The instance branch can be optimized by using cross entropy loss

$$\mathscr{L}_{inst} = \frac{1}{H\times W}\sum_{i}^{H\times W}\mathbb{1}_{\{i\in S_k\}}(-\log\varphi_k(c_{im},c_{in})) + \mathbb{1}_{\{i\in bg\}}(-\log(1 - \varphi_k(c_{im},c_{in}))). \tag{11}$$

$\varphi_k(c_{im},c_{in})$ is the same with Eq. (8). A smoothness term $\mathscr{L}_{smooth}$ is added to ensure that predicted $(\sigma_{im},\sigma_{in})$ is close to $(\sigma_{km},\sigma_{kn})$ in Eq. (10). In this way, the learned $(\sigma_{im},\sigma_{in})$ is stable for each pixel in the same instance and can be used as $(\hat{\sigma}_{km},\hat{\sigma}_{kn})$ during clustering in Eq. (10)

$$\mathscr{L}_{smooth} = \frac{1}{|S_k|}\sum_{i\in S_k}\|\sigma_{im}-\bar{\sigma}_{km}\|^2 + \|\sigma_{in}-\bar{\sigma}_{kn}\|^2. \tag{12}$$

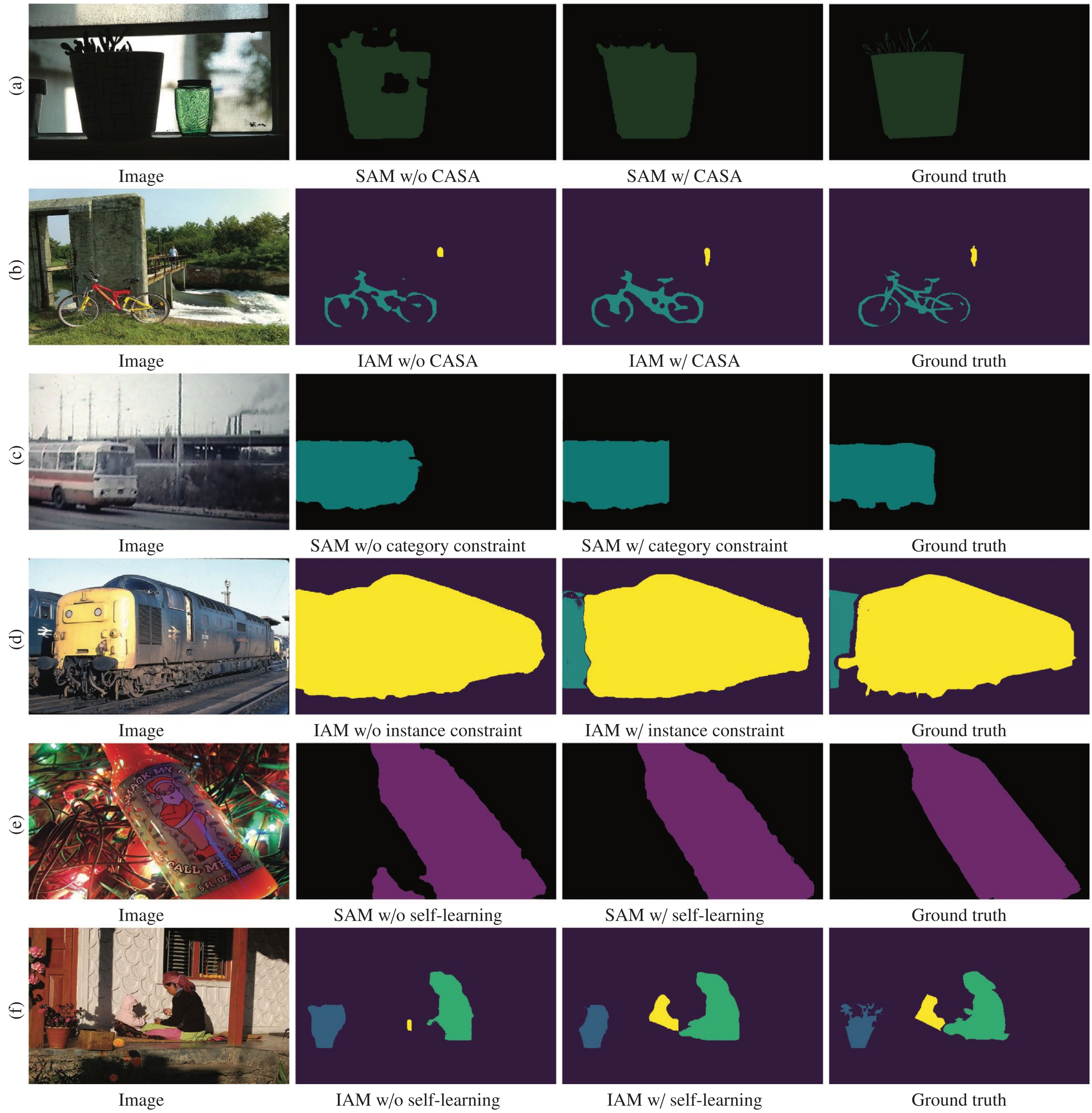

**Fig. 5.** Visualization on the PASCAL VOC 2012 [18]. We show results of semantic ancillary model (SAM) and instance ancillary model (IAM) with and without CASA, category/instance constraint and self-learning. Different colors indicate different categories in (a), (c) and (e), and indicate different instances in (b), (d) and (f).

## 4. Experimental Results

In this section, we first introduce datasets, metrics and implementation details in Section 4.1 and Section 4.2. Then we describe experimental results and ablation study in Section 4.4 and Section 4.3. Finally, we show results when our method extends to semantic segmentation in Section 4.5.

### 4.1. Datasets and Metrics

We demonstrate our method on two widely used datasets, *i.e.*, the PASCAL VOC 2012 [18] and the Cityscapes [14].

**PASCAL VOC 2012.** PASCAL VOC 2012 [18] and its augmented dataset [22] consist of 20 object categories and contains totally 10,582 training images [18,22], which is a common benchmark for semantic segmentation [44,12,61,53] and instance segmentation [23,30,64]. We follow [15,26,30,39,40,64] to adopt the mean average precision (*mAP*) for instance segmentation, and report *mAP* with three IoU (intersection over union) thresholds. They are 0.5, 0.7, and 0.75, denoted as $mAP^r_k$ where $k \in \{0.5, 0.7, 0.75\}$.

**Cityscapes.** The Cityscapes dataset [14] consists of 5,000 finely annotated images of $2048 \times 1024$ pixels, 2,975 for training set, 500 for validation set and 1,525 for testing set. The instance segmentation task consists of 8 different semantic categories. Notice that we only consider the "thing" category. It is a standard benchmark for

**Table 3**
Results on the PASCAL VOC 2012 [18] validation set. An asterisk "*" indicates using bounding boxes as pixel-wise labels for training.

| Supervision | Method | Backbone | Number of pixel-wise labels | Number of bounding box labels | Number of unlabeled images | Labeling cost | $mAP^r_{0.5}$ | $mAP^r_{0.7}$ | $mAP^r_{0.75}$ |
|---|---|---|---|---|---|---|---|---|---|
| Weak supervision | SDI (2017) [30] | VGG-16 | - | 10582 (100%) | - | 15.9% | 44.8 | - | 16.3 |
| | BBTP (2019) [26] | ResNet-101 | - | 10582 (100%) | - | 15.9% | 58.9 | 30.4 | 21.6 |
| | LACI (2020) [4] | ResNet-101 | - | 10582 (100%) | - | 15.9% | 57.7 | 33.5 | 31.2 |
| | *Mask R-CNN (2017) [24] | ResNet-101 | - | 10582 (100%) | - | 15.9% | 31.8 | 9.1 | 6.1 |
| | *CenterMask (2020) [38] | ResNeXt-152 | - | 10582 (100%) | - | 15.9% | 31.9 | 9.7 | 6.9 |
| Semi-supervision | Budget-aware (2019) [7] | ResNet-101 | 100 (0.9%) | - | 10482 (99.1%) | 0.9% | 14.9 | - | - |
| | | | 200 (1.9%) | - | 10382 (98.1%) | 1.9% | 23.7 | - | - |
| | | | 400 (3.8%) | - | 10182 (96.2%) | 3.8% | 35.5 | - | - |
| | | | 800 (7.6%) | - | 9782 (92.4%) | 7.6% | 42.9 | - | - |
| | | | 1464 (13.8%) | - | 9118 (86.2%) | 15.1% | 46.8 | - | - |
| Full supervision | SDS (2014) [23] | VGG-16 | 10582 (100%) | - | - | 100% | 49.7 | 25.3 | - |
| | MNC (2016) [15] | VGG-16 | 10582 (100%) | - | - | 100% | 63.5 | 41.5 | - |
| | SDI (2017) [30] | VGG-16 | 10582 (100%) | - | - | 100% | 47.5 | - | 20.2 |
| | SGN (2017) [43] | ResNet-101 | 10582 (100%) | - | - | 100% | 61.4 | 49.9 | - |
| | DIN (2017) [3] | ResNet-101 | 10582 (100%) | - | - | 100% | 61.7 | 48.6 | - |
| | FCIS (2017) [40] | ResNet-101 | 10582 (100%) | - | - | 100% | 65.7 | 52.1 | - |
| | RPE (2018) [35] | ResNet-101 | 10582 (100%) | - | - | 100% | 64.5 | - | - |
| | Budget-aware (2019) [7] | ResNet-101 | 10582 (100%) | - | - | 100% | 56.4 | - | - |
| | Mask R-CNN (2017) [24] | ResNet-101 | 105 (1%) | 10477 (99%) | - | 16.7% | 44.2 | 15.2 | 10.2 |
| | | | 317 (3%) | 10265 (97%) | - | 18.4% | 47.2 | 26.4 | 21.0 |
| | | | 529 (5%) | 10053 (95%) | - | 20.1% | 51.0 | 33.1 | 24.8 |
| | | | 1058 (10%) | 9524 (90%) | - | 24.3% | 54.9 | 38.4 | 31.6 |
| | | | 2116 (20%) | 8446 (80%) | - | 32.7% | 58.3 | 42.7 | 36.5 |
| | | | 5291 (50%) | 5291 (50%) | - | 57.9% | 63.8 | 47.5 | 39.9 |
| | | | 10582 (100%) | - | - | 100% | 67.9 | 50.7 | 43.5 |
| | CenterMask (2020) [38] | ResNeXt-152 | 105 (1%) | 10477 (99%) | - | 16.7% | 39.6 | 13.5 | 8.3 |
| | | | 317 (3%) | 10265 (97%) | - | 18.4% | 44.9 | 27.2 | 22.4 |
| | | | 529 (5%) | 10053 (95%) | - | 20.1% | 47.7 | 34.0 | 26.4 |
| | | | 1058 (10%) | 9524 (90%) | - | 24.3% | 54.2 | 39.5 | 34.6 |
| | | | 2116 (20%) | 8446 (80%) | - | 32.7% | 58.9 | 43.3 | 38.5 |
| | | | 5291 (50%) | 5291 (50%) | - | 57.9% | 63.9 | 49.0 | 42.8 |
| | | | 10582 (100%) | - | - | 100% | 68.8 | 53.1 | 45.9 |
| Hybrid supervision(Ours) | Mask R-CNN (2017) [24] | ResNet-101 | 105 (1%) | 10477 (99%) | - | 16.7% | 61.8 | 42.7 | 35.2 |
| | | | 317 (3%) | 10265 (97%) | - | 18.4% | 63.5 | 45.0 | 38.5 |
| | | | 529 (5%) | 10053 (95%) | - | 20.1% | 63.9 | 46.3 | 39.7 |
| | | | 1058 (10%) | 9524 (90%) | - | 24.3% | 65.4 | 48.0 | 41.6 |
| | | | 2116 (20%) | 8446 (80%) | - | 32.7% | 66.2 | 49.4 | 42.2 |
| | | | 5291 (50%) | 5291 (50%) | - | 57.9% | 66.9 | 50.0 | 43.5 |
| | CenterMask (2020) [38] | ResNeXt-152 | 105 (1%) | 10477 (99%) | - | 16.7% | 62.2 | 45.1 | 38.2 |
| | | | 317 (3%) | 10265 (97%) | - | 18.4% | 64.2 | 47.9 | 41.6 |
| | | | 529 (5%) | 10053 (95%) | - | 20.1% | 65.3 | 49.2 | 42.5 |
| | | | 1058 (10%) | 9524 (90%) | - | 24.3% | 66.1 | 50.8 | 44.6 |
| | | | 2116 (20%) | 8446 (80%) | - | 32.7% | 66.5 | 51.6 | 45.4 |
| | | | 5291 (50%) | 5291 (50%) | - | 57.9% | 67.2 | 52.1 | 45.9 |

**Table 4**
Results on the Cityscapes [14] validation set. An asterisk "*" indicates using bounding boxes as pixel-wise labels for training.

| Supervision | Method | Backbone | Number of pixel-wise labels | Number of bounding box labels | Labeling cost | $AP$ | $AP_{50}$ |
|---|---|---|---|---|---|---|---|
| Weak supervision | WSSPS (2018) [39] | ResNet-101 | - | 2975 (100%) | 9.2% | 17.0 | - |
| | *Mask R-CNN (2017) [24] | ResNet-101 | - | 2975 (100%) | 9.2% | 12.0 | 35.3 |
| | *CenterMask (2020) [38] | ResNeXt-152 | - | 2975 (100%) | 9.2% | 12.3 | 34.0 |
| Full supervision | InstanceCut (2017) [34] | ResNet-101 | 2975 (100%) | - | 100% | 15.8 | 32.4 |
| | DWT (2017) [5] | VGG-16 | 2975 (100%) | - | 100% | 21.2 | - |
| | SGN (2017) [43] | ResNet-101 | 2975 (100%) | - | 100% | 29.2 | 49.4 |
| | WSSPS (2018) [39] | ResNet-101 | 2975 (100%) | - | 100% | 24.3 | - |
| | Mask R-CNN (2017) [24] | ResNet-50 | 29 (1%) | 2946 (99%) | 10.1% | 15.1 | 39.8 |
| | | | 89 (3%) | 2886 (97%) | 11.9% | 16.3 | 45.1 |
| | | | 148 (5%) | 2827 (95%) | 13.7% | 18.7 | 48.1 |
| | | | 297 (10%) | 2678 (90%) | 18.3% | 23.4 | 51.0 |
| | | | 595 (20%) | 2380 (80%) | 27.4% | 26.7 | 53.7 |
| | | | 1487 (50%) | 1488 (50%) | 54.6% | 30.1 | 56.9 |
| | | | 2975 (100%) | - | 100% | 31.5 | 59.5 |
| | CenterMask (2020) [38] | ResNeXt-152 | 29 (1%) | 2946 (99%) | 10.1% | 13.3 | 39.1 |
| | | | 89 (3%) | 2886 (97%) | 11.9% | 16.6 | 46.2 |
| | | | 148 (5%) | 2827 (95%) | 13.7% | 20.5 | 49.5 |
| | | | 297 (10%) | 2678 (90%) | 18.3% | 27.1 | 54.3 |
| | | | 595 (20%) | 2380 (80%) | 27.4% | 29.6 | 56.7 |
| | | | 1487 (50%) | 1488 (50%) | 54.6% | 32.9 | 59.5 |
| | | | 2975 (100%) | - | 100% | 34.7 | 61.7 |
| Hybrid supervision(Ours) | Mask R-CNN (2017) [24] | ResNet-50 | 29 (1%) | 2946 (99%) | 10.1% | 19.4 | 40.2 |
| | | | 89 (3%) | 2886 (97%) | 11.9% | 24.2 | 49.6 |
| | | | 148 (5%) | 2827 (95%) | 13.7% | 25.8 | 51.8 |
| | | | 297 (10%) | 2678 (90%) | 18.3% | 28.5 | 55.0 |
| | | | 595 (20%) | 2380 (80%) | 27.4% | 29.5 | 56.8 |
| | | | 1487 (50%) | 1488 (50%) | 54.6% | 31.1 | 59.2 |
| | CenterMask (2020) [38] | ResNeXt-152 | 29 (1%) | 2946 (99%) | 10.1% | 18.7 | 39.9 |
| | | | 89 (3%) | 2886 (97%) | 11.9% | 26.7 | 51.8 |
| | | | 148 (5%) | 2827 (95%) | 13.7% | 27.4 | 54.0 |
| | | | 297 (10%) | 2678 (90%) | 18.3% | 30.5 | 56.1 |
| | | | 595 (20%) | 2380 (80%) | 27.4% | 32.1 | 57.9 |
| | | | 1487 (50%) | 1488 (50%) | 54.6% | 34.0 | 61.4 |

**Table 5**
Results of the semantic ancillary model on the PASCAL VOC 2012 [18] validation set.

| CASA | Category constraint | Self-learning | *mIoU* |
|---|---|---|---|
| × | × | × | 76.6 |
| √ | × | × | 86.5 |
| √ | √ | × | 89.1 |
| √ | √ | √ | **90.2** |

instance segmentation in driving scenes. We measure the performance by using COCO-style mask $AP$ (averaged over thresholds from 0.5 to 0.95 with an interval of 0.05) and $AP_{50}$ (*i.e.*, mask AP at an IoU of 0.5), following [24,43,5,39].

### 4.2. Implementation Details

Here, we introduce the implementation details of our semantic ancillary model, instance ancillary model, and primary instance segmentation models on the both PASCAL VOC 2012 [18] and Cityscapes [14]. Notice that bounding box labels can be obtained from pixel-wise labels by finding extreme points.

**Training of semantic ancillary model.** In current implementation, the backbone network is based on Xception-65 [12] and ADAM optimizer [31] is utilized. It is pre-trained on the ImageNet dataset [51]. We use ADAM optimizer [31] and polynomial learning rate decay $(1 - \frac{iter}{max\ iter})^{0.9}$. The initial learning rate and batch size are set to $7e-4$ and 8. The cropping size is set to $512 \times 512$ on the PASCAL VOC 2012 [18] and $640 \times 640$ on the Cityscapes [14]. The cross entropy loss and lovász loss [8] are combined as its loss function. We train the model for 60 k iterations with four GeForce GTX 1080 Ti GPUs.

**Training of instance ancillary model.** In the current implementation, we also use the same backbone network, optimizer, and polynomial learning rate decay as those in the semantic ancillary model. The spatial-embedding loss [45] is used as the loss function. We use offset value between $[-1, 1]$ to represent $[-512, 512]$ and $[-1024, 1024]$ pixel offset in the PASCAL VOC [18] and Cityscapes [14] respectively.

**Table 6**
Ablation studies on the PASCAL VOC 2012 [18] validation set. To verify the effectiveness of each module, we compare generated pseudo labels with ground truth to evaluate their quality.

| Instance ancillary model | CASA | Semantic segmentation result | Instance constraint | Self-Learning | $mAP^r_{0.5}$ | $mAP^r_{0.75}$ |
|---|---|---|---|---|---|---|
| × | × | × | × | × | 21.6 | 1.8 |
| √ | × | × | × | × | 47.6 | 28.4 |
| √ | √ | × | × | × | 63.8 | 38.4 |
| √ | √ | √ | × | × | 65.6 | 43.3 |
| √ | √ | √ | √ | × | 70.1 | 47.8 |
| √ | √ | √ | √ | √ | **70.2** | **49.3** |

**Table 7**
Ablation study for self-learning on the PASCAL VOC 2012 [18] validation set. Self-learning is proposed to solve the overfitting problem and can significantly improve the quality of the pseudo labels when the number of pixel-wise labeled samples is very small.

| Self-learning | Number of pixel-wise labels | Number of bounding box labels | $mAP^r_{0.5}$ | $mAP^r_{0.75}$ |
|---|---|---|---|---|
| × | 105(1%) | 10477(99%) | 38.2 | 11.8 |
| √ | 105(1%) | 10477(99%) | **64.2** | **39.1** |
| × | 1058(10%) | 9524(90%) | 70.1 | 47.8 |
| √ | 1058(10%) | 9524(90%) | **70.2** | **49.3** |

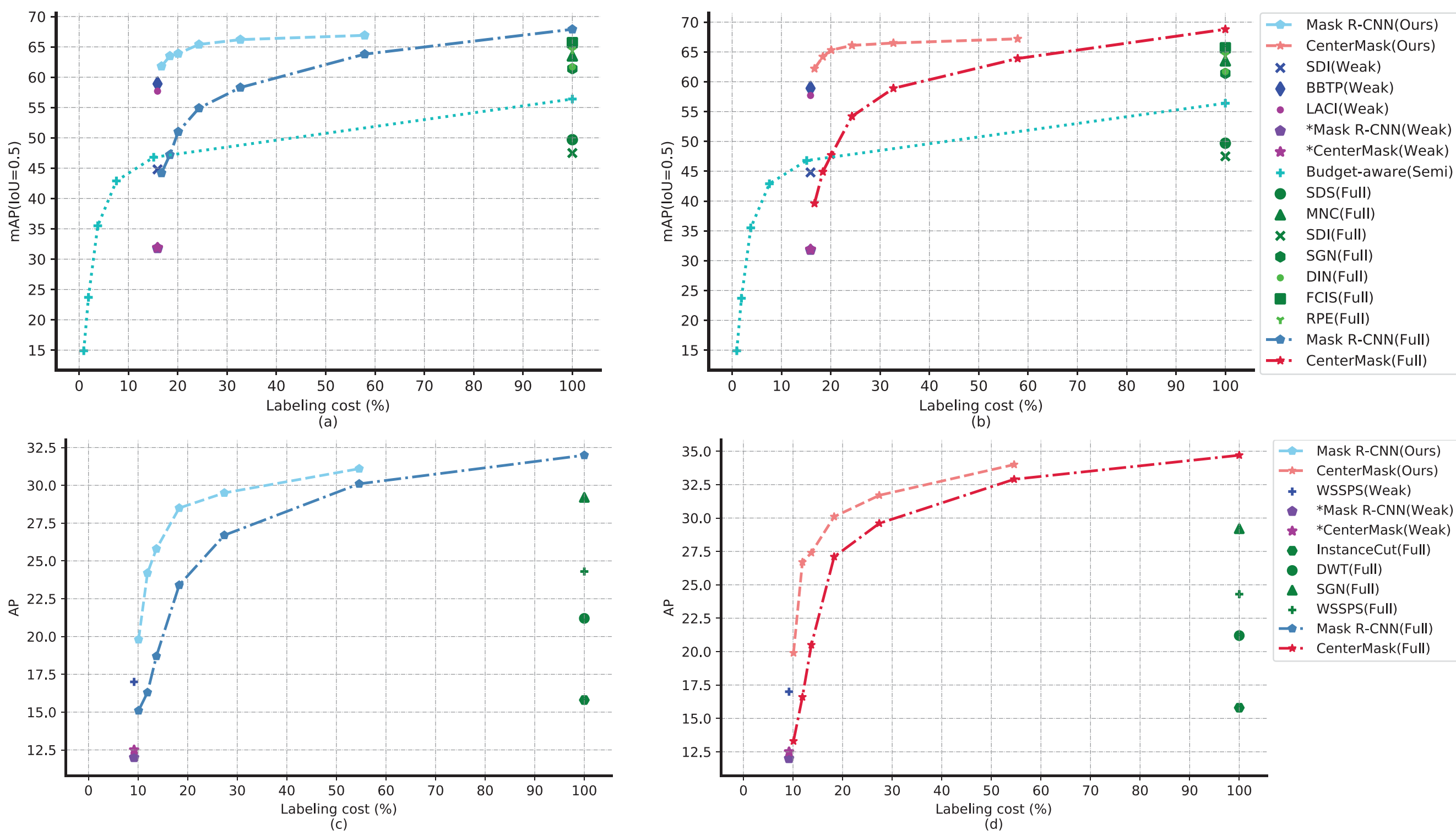

**Fig. 6.** Evaluation of instance segmentation results from different methods. (a), (b) are $mAP^r_{0.5}$ results Mask R-CNN [24] and CenterMask [38] on the PASCAL VOC 2012 [18]. (c), (d) are *AP* results of Mask R-CNN [24] and CenterMask [38] on the Cityscapes [14]. The horizontal axis is the percentage of pixel-wise labels in dataset. The vertical axis is performance metric.

For the PASCAL VOC 2012 [18], the initial learning rate is $1e-4$. We train the model on $448 \times 448$ crops for 60 k iterations with batch size of 8. For the Cityscapes [14], following [45], the initial learning rate is $5e-4$, which is decreased to $5e-5$ for finetuning. We first pre-train our model for 40 k iterations with a batch size of 8 on $448 \times 448$ crops that are centered around an object to avoid spending too much computation time on background patches without any instances. Then, we finetune the model for another 20 k iterations on $864 \times 864$ crops with a batch size of 2, and fix the batch normalization statistics during this stage. We follow [45] to use Lovász-hinge loss [8] instead of cross entropy in Eq. (11) to maximize the intersection-over-union for each instance mask. The instance ancillary model is optimized with two GeForce GTX 1080 Ti GPUs.

**Training of primary instance segmentation model.** Our proposed method can generally adopt any instance segmentation model as primary instance segmentation model. Here, we choose two widely used and recent networks, *i.e.*, Mask R-CNN [24] and CenterMask [38]. For the training setting, batch size, learning rate, weight decay, momentum, and the number of the iterations are set to 8, $1e-2$, $1e-4$, 0.9 and 24 k, respectively. The ADAM optimizer [31] is utilized.

In our implementation, Mask R-CNN [24] adopts ResNet-101 [25] as backbone network on the PASCAL VOC 2012 [18]. For the Cityscapes [14], we follow [24] and use ResNet-50 [25] as the backbone network. ResNet-101 and ResNet-50 [25] are pre-trained on ImageNet dataset [51]. For CenterMask [38], we choose ResNeXt-152 [57] as backbone network for all datasets. ResNeXt-152 [57] is pre-trained on the ImageNet dataset [51]. The initial learning rate is $1e-2$ and drops to $1e-3$ at the 16 k-th iterations on the PASCAL VOC [18] and at the 18 k-th iterations on the Cityscapes [14]. After 24 k iterations, we finetune the model for another 2 k iterations with pixel-wise labeled images. For data augmentation, we use the same settings as those in Mask R-CNN [24]. Both Mask R-CNN [24] and CenterMask [38] are optimized on eight GeForce GTX 1080 Ti GPUs.

### 4.3. Ablation Study

In this section, we conduct ablation studies on category-aware spatial attention module, category constraint module, instance constraint module, and self-learning to investigate how they improve the label noise suppression ability of ancillary models.

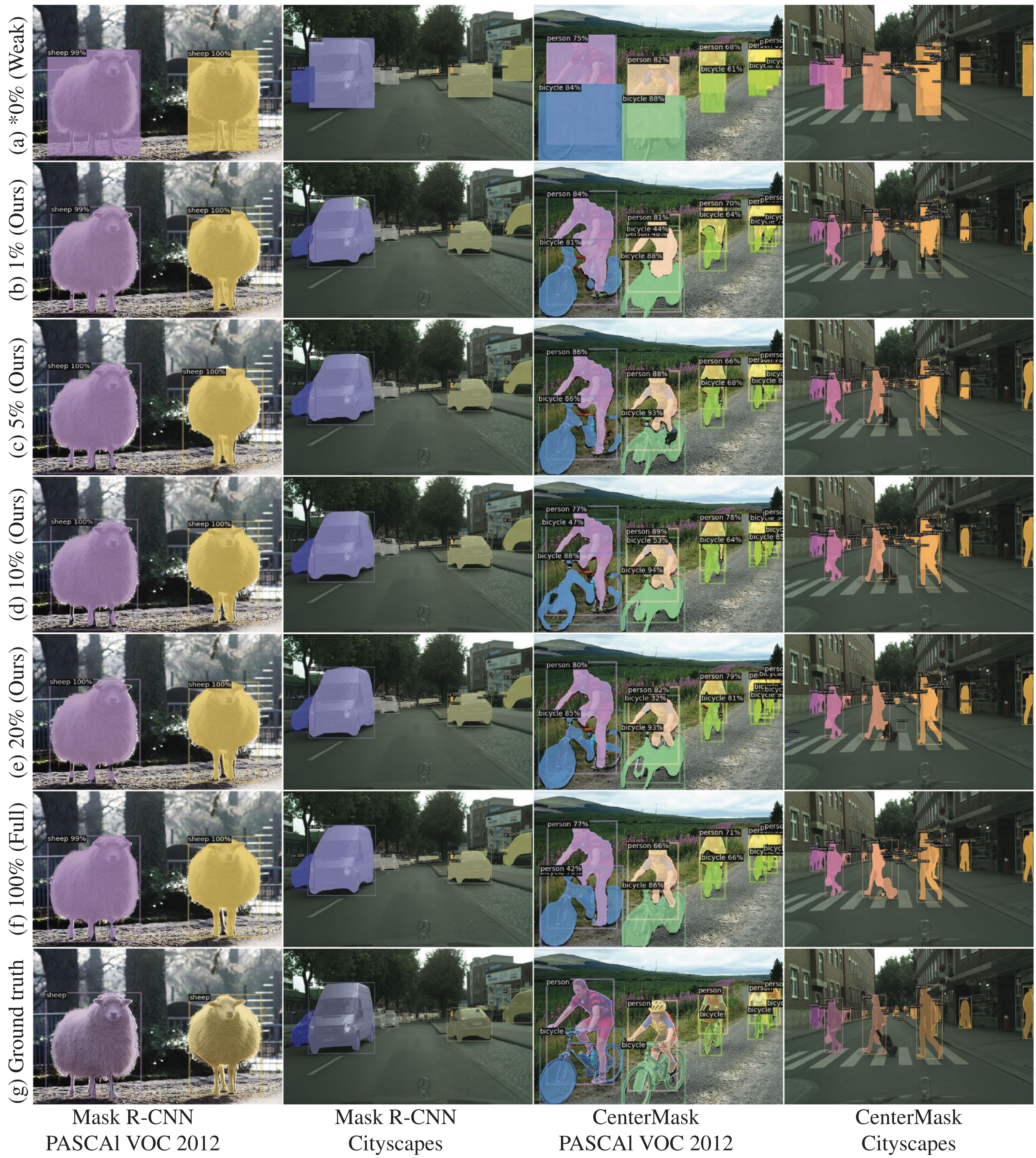

**Fig. 7.** Representative visualized results of Mask R-CNN [24] and CenterMask [38] with different supervised settings on the PASCAl VOC 2012 [18] and Cityscapes [14]. Our results are basically the same with a fully supervised setting and visually very close to the .ground truth.

We quantify their contributions with *mIoU* [12], $mAP^r_{0.5}$, and $mAP^r_{0.75}$ [30]. Higher *mIoU* indicates lower background noise level, and higher *mAP* indicates both lower background noise level and lower instance noise level. Experiments are conducted on the augmented PASCAL VOC 2012 dataset [18,22] with 10% pixel-wise labels. Results are provided in Table 5 and Table 6. To further show the effectiveness of self-learning, we additionally provide results with 1% pixel-wise labels in Table 7.

**Category-aware spatial attention (CASA) module.** As shown in Table 5 and Table 6, the CASA module significantly increases the *mIoU* from 76.6 to 86.5 and improves the $mAP^r_{0.5}$ and $mAP^r_{0.75}$ by 16.2 and 10.0 respectively. These show it can help ancillary models to suppress the background noise and instance noise effectively.

**Semantic segmentation results.** Though the instance ancillary model is able to suppress both background noise and instance noise in bounding box labels independently, it can be seen in

**Table 8**
Results of the semantic segmentation on the PASCAL VOC 2012 [17] validation set. Notice that we use flipping and multi-scale processing for images like in [12], and we follow [7] to assume that semantic segmentation label requires the same labeling cost as instance segmentation label does.

| Method | Backbone | Number of pixel-wise labels | Number of bounding box labels | *mIoU* |
|---|---|---|---|---|
| Song et al. [52] | ResNet-101 [25] | 1464(13.8%) | 9118(86.2%) | 71.60 % |
| Ibrahim et al. [29] | Xecption-65 [12] | 1464(13.8%) | 9118(86.2%) | 82.33 % |
| Vanilla DeepLabv3+ [12] | Xecption-65 [12] | 10582(100%) | - | 81.21 % |
| Deeplabv3+ (Ours) | Xecption-65 [12] | 105(1%) | 10477(99%) | 79.13 |
| | | 317(3%) | 10265(99%) | 79.72% |
| | | 529(5%) | 10053(95%) | 81.06% |
| | | 1058(10%) | 9524(90%) | 83.28% |
| | | 1464(13.8%) | 9118(86.2%) | **83.41**% |

Table 6 that the semantic segmentation model dramatically improves the $mAP^r_{0.75}$ from 38.4 to 43.3. This means that the semantic segmentation model helps to suppress the background noise and ease the burden on the instance ancillary model, so the instance ancillary model can focus on suppressing the instance noise and predicting better pseudo labels.

**Category constraint module** & **Instance constraint module.** Table 5 shows category constraint module increase the *mIoU* from 86.5 to 89.1, and instance constraint module greatly increases the $mAP^r_{0.5}$ and $mAP^r_{0.75}$ by 5.4 and 4.9, as illustrated in Table 6. They effectively refine the pseudo labels by keeping consistency between bounding box labels and pseudo labels.

**Self-learning.** Here, we first conduct an ablation study for self-learning with 10% pixel-wise labeled samples (1058 samples). The number of samples is relatively large and the overfitting problem is not so serious. Under this setting, self-learning contributes to extra 1.1 improvement on *mIoU* as shown in Table 5 and 1.5 improvement on $mAP^r_{0.75}$ as shown in Table 6. When we reduce pixel-wise labels to 1% (105 samples), the overfitting problem becomes serious and the quality of generated pseudo labels is largely degraded as shown in Table 7. By utilizing pseudo labels generated from bounding box labeled samples, the self-learning training approach can solve the overfitting problem and significantly improve the $mAP^r_{75}$ by 27.3. In short, self-learning improves the robustness of the proposed method when the number of pixel-wise labeled samples is very small.

### 4.4. Comparison with State-of-the-art Methods

We conduct quantitative evaluation with different percentage of pixel-wise labels and bounding box labels. We test our method on two different primary instance segmentation models[1]: two-stage Mask R-CNN [24] and one-stage CenterMask [38]. The results of compared methods are from their original papers[2]. For a fair comparison, we unify the analysis across different supervision settings by considering labeling cost (the lower is the better) and performance (the higher is the better). The labeling cost is calculated using statistics in Table 1. And for a comparative purpose, it is represented as a percentage compared with full supervision in Table 3 and Table 4. The overall results are illustrated in Fig. 6, the proposed method achieves much better performance than fully supervised baseline under the various percent of pixel-wise labels (*i.e.*, 1%, 3%, 5%, 10%, 20% and 50%), which shows the effectiveness of the proposed method. Notice that the bounding box labels are only used for training the bounding box regression branch and classification branch of the model in a fully supervised setting.

For the PASCAL VOC 2012 [18], we compare the proposed method with fifteen state-of-the-art methods, including ten fully supervised methods (*i.e.*, SDS [23], MNC [15], SDI [30], SGN [43], DIN [3], FCIS [40], RPE [35], Budget-aware [7], Mask R-CNN [24] and CenterMask [38]), one semi-supervised method (*i.e.*, Budget-aware [7]), and five weakly supervised methods (*i.e.*, SDI [30], BBTP [26], LACI [4], Mask R-CNN [24] and CenterMask [38] trained with only bounding box labels). The detailed results for all compared methods are illustrated in Table 3.

Taking the Mask R-CNN [24] based implementation as an example, we first analyze the results with 1% pixel-wise labels. It can be seen that the proposed method trained with 1% pixel-wise labels outperforms semi-supervised Budget-aware [7] with 13.8% pixel-wise labels by 15.0 in $mAP^r_{0.5}$. For $mAP^r_{0.75}$, the proposed method surpasses weakly supervised method SDI [30], BBTP [26] and LACI [4] by large margins, around 18.9, 13.6 and 4.0 respectively. Notice that the labeling cost is similar, *i.e.*, Budget-aware [7] 15.1%, SDI [30], BBTP [26] and LACI [4] 15.9%, ours 16.7%. Besides, it outperforms fully supervised SDS [23], MNC [15] by 17.4 and 1.2 in $mAP^r_{0.7}$, and achieves comparable results to fully supervised SGN [43], DIN [3], FCIS [40] and RPE [35].

We further analyze the results with 10% pixel-wise labels. The proposed method achieves comparable accuracy with fully supervised FCIS [40], Mask R-CNN [24], CenterMask [38]. It outperforms SDS [23], MNC [15] by 22.7, 6.5 in $mAP^r_{0.7}$, and achieves 4.0, 3.7, 0.9 better results than SGN [43], DIN [3], RPE [35] in $mAP^r_{0.5}$. If we switch the primary instance segmentation to CenterMask [38] with a stronger backbone [57] in our implementation, the results are even better.Moreover, with 50% pixel-wise labels, our Mask R-CNN [24] and CenterMask [38] achieve the same $mAP^r_{0.75}$ results with full supervision, which means the proposed method keeps similar performance while reducing 42.1% labeling cost.

As for the Cityscapes [18], we compare the proposed method with nine state-of-the-art methods including six fully supervised methods (*i.e.*, InstanceCut [34], DWT [5], SGN [43], WSSPS [39], Mask R-CNN [24] and CenterMask [38]), and three weakly supervised methods (*i.e.*, WSSPS [39], Mask R-CNN [24] and CenterMask [38] trained with only bounding box labels). The detailed results are provided in Table 4.

For Mask R-CNN [24] as primary instance segmentation model, the proposed method using 1% pixel-wise labels (*i.e.*, 29 samples) performs better than weakly supervised WSSPS [39], *i.e.*, 17.0 *AP* *vs.* 19.4 *AP*, as shown in Table 4. Notice that WSSPS [39] has stronger backbone than our Mask R-CNN [24], *i.e.*, ResNet-101 [25] *vs.* ResNet-50 [25], while labeling cost is similar, *i.e.*, WSSPS [39] 9.2%, ours 10.1%. And with 10% pixel-wise labels, it performs comparable to fully supervised SGN [43], Mask R-CNN [24], CenterMask [38] and better than fully supervised InstanceCut [34], DWT [5], WSSPS [39]. Similarly, when we switch the primary instance segmentation model to CenterMask [38] with a stronger backbone [57], the results are even better.

Fig. 7 shows representative examples. We can see that as the percentage of pixel-wise labels increases, the quality of predictions improves. The proposed method with 1% pixel-wise labels predicts much better than the 0% setting (weak supervision), and is comparable to the fully supervised setting. When the percentage

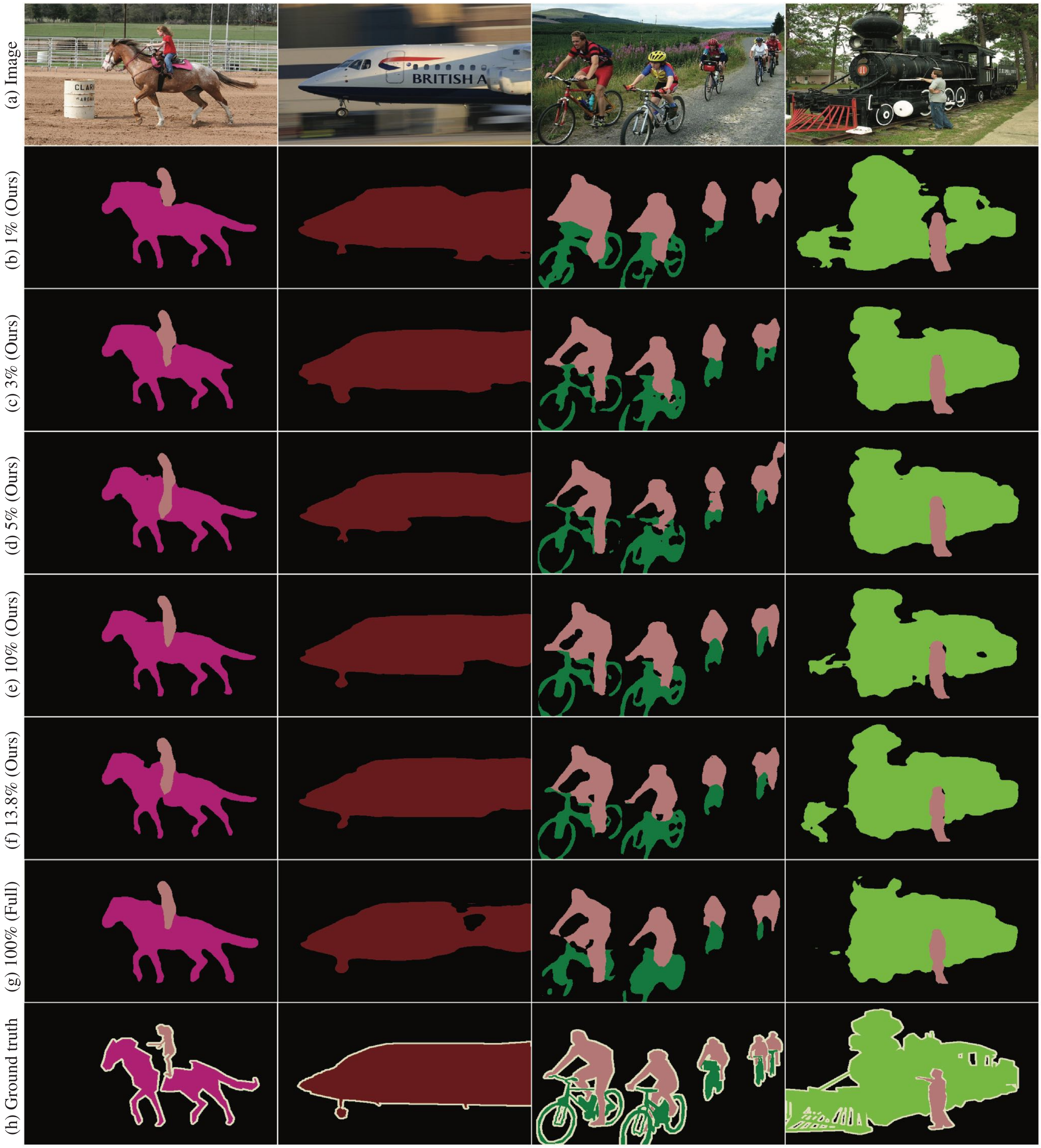

**Fig. 8.** We compare our method (b)-(f) with fully supervised DeepLabv3+ [12] (g) on the PASCAL VOC 2012 [18] validation set. Our method performs similarly or even better than fully supervised DeepLabv3+ [12] when trained with 10% or more pixel.-wise labels.

increases to 10%, the visual quality is already very close to the full supervision.

### 4.5. Extension to Semantic Segmentation

Semantic segmentation is a similar task. The proposed method can generally apply to it by removing the instance ancillary model and replacing the primary instance segmentation model with the primary semantic segmentation model. Particularly, we use the DeepLabv3+ [12] as the primary semantic segmentation model.

Generally, we can first train the semantic ancillary model and predict pseudo labels, then optimize the primary DeepLabv3+ [12] with these pseudo labels. We notice the semantic ancillary model can be seen as a DeepLabv3+ [12] with category-aware spatial attention module and category constraint module, so we let the primary DeepLabv3+ [12] share the parameters with semantic ancillary model and optimize them jointly in an end-to-end way.

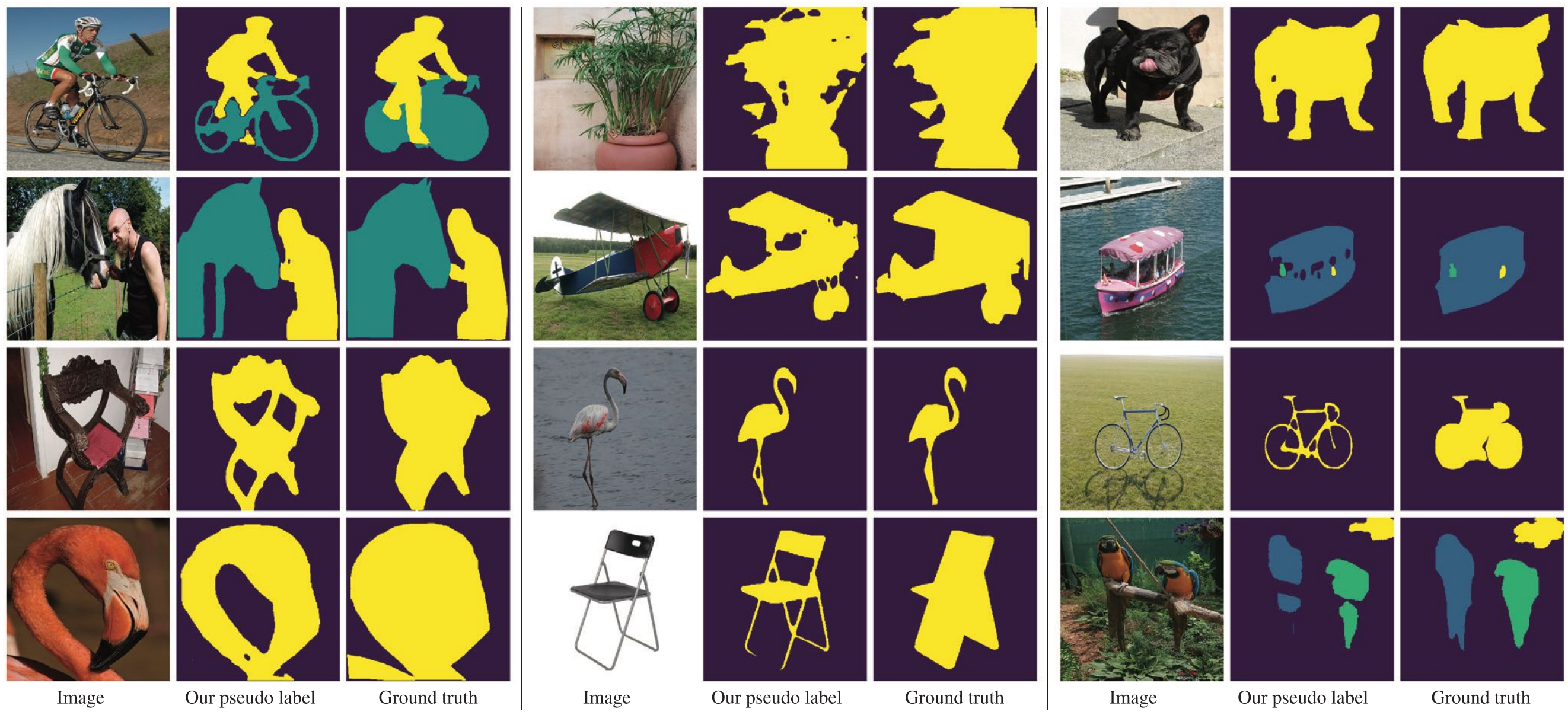

**Fig. 9.** Comparison between our pseudo instance-level pixel-wise labels and noisy ground truth on SBD [22] dataset. Different colors indicate different instances. The pipeline is trained with 10% pixel-wise labels from the PASCAL VOC 2012 [18].

During the training of each epoch, we first train the semantic ancillary model and DeepLabv3+ [12] with ground truth pixel-wise labels, then we train the primary DeepLabv3+ [12] with pseudo pixel-wise labels produced by the semantic ancillary model. Notice that we ignore losses from pixels of uncertain categories (whose max confidence values are less than 0.5 in pseudo labels). Since the primary DeepLabv3+ [12] shares the parameters with the semantic ancillary model, the semantic ancillary model is also updated. And in the next epoch, the updated semantic ancillary model can predict more accurate pseudo pixel-wise labels to further improve the primary DeepLabv3+ [12]. In this way, the semantic ancillary model is able to predict more and more accurate pseudo labels and the primary DeepLabv3+ [12] can achieve competitive results after training.

Following [44,12], we evaluate the performance using *mIoU* metric on the PASCAL VOC 2012 [18]. As shown in Table 8, when the labeling cost is the same, our method performs better than Ibrahim et al. [29]. Notice that our model can be trained in an end-to-end way, while Ibrahim et al. [29] cannot. As shown in Table 8 and Fig. 8, we can see that our method even outperforms the fully supervised setting both quantitatively and qualitatively with only 10% pixel-wise labels, which means the proposed method can achieve 2.2 better *mIoU* with more than 75% labeling cost reduction. The reason may be that the fully supervised method is sensitive to label noise coming from its augmented training set [22].

## 5. Conclusion

In this paper, we present a pipeline of hybrid supervision that can fully utilize pixel-wise labels and bounding box labels, it well balances performance and labeling cost for instance segmentation. Specifically, we design a semantic ancillary model and an instance ancillary model to suppress the background noise and instance noise in bounding box labels. Moreover, we exploit category-aware spatial attention module, category constraint module, instance constraint module to fully utilize bounding box labels, so as to produce high-quality pseudo instance-level pixel-wise labels. And we propose a self-learning training approach to avoid the overfitting problem. Experimental results on the PASCAL VOC 2012 [18] and the Cityscapes [14] show that our method not only outperforms weakly and semi-supervised methods [30,39,26,4,7] by a large margin, but also achieves better performance than some former state-of-the-art fully supervised methods [23,15,34,5,43,3,35] with much less labeling cost.Our pipeline is able to learn the accurate shapes of objects from original PASCAL VOC 2012 [18], suppress the noise, and even generate pseudo labels better than noisy ground truth in [22], as shown in Fig. 9. We also extend our method to the semantic segmentation task, and achieve better performance than full supervision on the PASCAL VOC 2012 [18] with more than 75% labeling cost reduction.

Our work mainly focuses on designing ancillary models to produce accurate pseudo pixel-wise labels for instance/semantic segmentation tasks and has not designed the primary segmentation model. In the future, it is worth investigating how to design a primary segmentation model that is robust to noise in labels, and further extend our work to other pixel-wise prediction tasks *e.g.*, panoptic segmentation [33,32].

## CRediT authorship contribution statement

**Linwei Chen:** Conceptualization, Methodology, Software, Writing - original draft, Visualization, Investigation. **Ying Fu:** Writing - review & editing, Supervision. **Shaodi You:** Writing - review & editing. **Hongzhe Liu:** Resources.

## Declaration of Competing Interest

The authors declare that they have no known competing financial interests or personal relationships that could have appeared to influence the work reported in this paper.

## Acknowledgment

This work was supported by the National Natural Science Foundation of China under Grants No. 62171038, No. 61827901, No. 61936011, and No. 62088101.

## References

[1] Jiwoon Ahn, Sunghyun Cho, Su.ha. Kwak, Weakly supervised learning of instance segmentation with inter-pixel relations, in: Proceedings of IEEE International Conference on Computer Vision, 2019, pp. 2209–2218.

[2] Pablo Arbeláez, Jordi Pont-Tuset, Jonathan T Barron, Ferran Marques, and Jitendra Malik. Multiscale combinatorial grouping. In Proceedings of IEEE International Conference on Computer Vision, pages 328–335, 2014.

[3] Anurag Arnab, Philip HS Torr, Pixelwise instance segmentation with a dynamically instantiated network, in: Proceedings of IEEE Conference on Computer Vision and Pattern Recognition, 2017, pp. 441–450.

[4] Aditya Arun, CV Jawahar, and M Pawan Kumar. Weakly supervised instance segmentation by learning annotation consistent instances. In Proceedings of European Conference on Computer Vision, pages 254–270. Springer, 2020.

[5] Min Bai, Raquel Urtasun, Deep watershed transform for instance segmentation, in: Proceedings of IEEE Conference on Computer Vision and Pattern Recognition, 2017, pp. 5221–5229.

[6] Amy Bearman, Olga Russakovsky, Vittorio Ferrari, Li Fei-Fei, What's the point: Semantic segmentation with point supervision, in: Proceedings of European Conference on Computer Vision, 2016, pp. 549–565.

[7] Míriam Bellver Bueno, Amaia Salvador Aguilera, Jordi Torres Viñals, and Xavier Giró Nieto. Budget-aware semi-supervised semantic and instance segmentation. In Proceedings of IEEE Conference on Computer Vision and Pattern Recognition Workshops, pages 93–102, 2019.

[8] Maxim Berman, Amal Rannen Triki, Matthew B Blaschko, The lovász-softmax loss: A tractable surrogate for the optimization of the intersection-over-union measure in neural networks, in: Proceedings of IEEE International Conference on Computer Vision, 2018, pp. 4413–4421.

[9] Daniel Bolya, Chong Zhou, Fanyi Xiao, and Yong Jae Lee. Yolact: real-time instance segmentation, in: Proceedings of IEEE International Conference on Computer Vision, 2019, pp. 9157–9166.

[10] Hao Chen, Kunyang Sun, Zhi Tian, Chunhua Shen, Yongming Huang, and Youliang Yan. Blendmask: Top-down meets bottom-up for instance segmentation. In Proceedings of IEEE Conference on Computer Vision and Pattern Recognition, pages 8573–8581, 2020.

[11] Kai Chen, Jiangmiao Pang, Jiaqi Wang, Yu Xiong, Xiaoxiao Li, Shuyang Sun, Wansen Feng, Ziwei Liu, Jianping Shi, Wanli Ouyang, et al. Hybrid task cascade for instance segmentation. In Proceedings of IEEE International Conference on Computer Vision, pages 4974–4983, 2019.

[12] Liang-Chieh Chen, Yukun Zhu, George Papandreou, Florian Schroff, and Hartwig Adam. Encoder-decoder with atrous separable convolution for semantic image segmentation. In Proceedings of European Conference on Computer Vision, pages 801–818, 2018.

[13] Hisham Cholakkal, Guolei Sun, Fahad Shahbaz Khan, Ling Shao, Object counting and instance segmentation with image-level supervision, in: Proceedings of IEEE Conference on Computer Vision and Pattern Recognition, 2019, pp. 12397–12405.

[14] Marius Cordts, Mohamed Omran, Sebastian Ramos, Timo Rehfeld, Markus Enzweiler, Rodrigo Benenson, Uwe Franke, Stefan Roth, and Bernt Schiele. The cityscapes dataset for semantic urban scene understanding. In Proceedings of IEEE International Conference on Computer Vision, pages 3213–3223, 2016.

[15] Jifeng Dai, Kaiming He, Jian Sun, Instance-aware semantic segmentation via multi-task network cascades, in: Proceedings of IEEE Conference on Computer Vision and Pattern Recognition, 2016, pp. 3150–3158.

[16] Bert De Brabandere, Davy Neven, Luc Van Gool, Semantic instance segmentation for autonomous driving, in: Proceedings of IEEE Conference on Computer Vision and Pattern Recognition Workshops, 2017, pp. 7–9.

[17] S.M. Mark Everingham, Ali Eslami, Luc Van Gool, Christopher KI Williams, John Winn, Andrew Zisserman, The pascal visual object classes challenge: A retrospective, International Journal of Computer Vision 111 (1) (2015) 98–136.

[18] Mark Everingham, Luc Van Gool, Christopher KI Williams, John Winn, and Andrew Zisserman. The pascal visual object classes (voc) challenge. International Journal of Computer Vision, 88(2), 303–338, 2010.

[19] Kuan Fang, Yunfei Bai, Stefan Hinterstoisser, Silvio Savarese, Mrinal Kalakrishnan, Multi-task domain adaptation for deep learning of instance grasping from simulation, in: Proceedings of IEEE International Conference on Robotics and Automation, 2018, pp. 3516–3523.

[20] Jun Fu, Jing Liu, Haijie Tian, Yong Li, Yongjun Bao, Zhiwei Fang, and Hanqing Lu. Dual attention network for scene segmentation. In Proceedings of IEEE International Conference on Computer Vision, pages 3146–3154, 2019.

[21] Weifeng Ge, Sheng Guo, Weilin Huang, Matthew R Scott, Label-penet: Sequential label propagation and enhancement networks for weakly supervised instance segmentation, in: Proceedings of IEEE International Conference on Computer Vision, 2019, pp. 3345–3354.

[22] Bharath Hariharan, Pablo Arbeláez, Lubomir Bourdev, Subhransu Maji, Jitendra Malik, Semantic contours from inverse detectors, in: Proceedings of IEEE International Conference on Computer Vision, 2011, pp. 991–998.

[23] Bharath Hariharan, Pablo Arbeláez, Ross Girshick, Jitendra Malik, Simultaneous detection and segmentation, in: Proceedings of European Conference on Computer Vision, 2014, pp. 297–312.

[24] Kaiming He, Georgia Gkioxari, Piotr Dollár, Ross Girshick, Mask r-cnn, in: Proceedings of IEEE International Conference on Computer Vision, 2017, pp. 2961–2969.

[25] Kaiming He, Xiangyu Zhang, Shaoqing Ren, Jian Sun, Deep residual learning for image recognition, in: Proceedings of IEEE International Conference on Computer Vision, 2016, pp. 770–778.

[26] Cheng-Chun Hsu, Kuang-Jui Hsu, Chung-Chi Tsai, Yen-Yu Lin, and Yung-Yu Chuang. Weakly supervised instance segmentation using the bounding box tightness prior, in: Proceedings of Advances in Neural Information Processing Systems, 2019, pp. 6586–6597.

[27] Hu. Ronghang, Piotr Dollár, Kaiming He, Trevor Darrell, Ross Girshick, Learning to segment every thing, in: Proceedings of IEEE International Conference on Computer Vision, 2018, pp. 4233–4241.

[28] Wei-Chih Hung, Yi-Hsuan Tsai, Xiaohui Shen, Zhe Lin, Kalyan Sunkavalli, Xin Lu, and Ming-Hsuan Yang. Scene parsing with global context embedding. In Proceedings of IEEE International Conference on Computer Vision, pages 2631–2639, 2017.

[29] Mostafa S Ibrahim, Arash Vahdat, and William G Macready. Semi-supervised semantic image segmentation with self-correcting networks. pages 12715–12725, 2020.

[30] Anna Khoreva, Rodrigo Benenson, Jan Hosang, Matthias Hein, Bernt Schiele, Simple does it: Weakly supervised instance and semantic segmentation, in: Proceedings of IEEE International Conference on Computer Vision, 2017, pp. 876–885.

[31] Diederik P Kingma, Jimmy Ba, Adam: A method for stochastic optimization, in: Proceedings of International Conference on Learning Representations, 2014, pp. 1–15.

[32] Alexander Kirillov, Ross Girshick, Kaiming He, Piotr Dollár, Panoptic feature pyramid networks, in: Proceedings of IEEE Conference on Computer Vision and Pattern Recognition, 2019, pp. 6399–6408.

[33] Alexander Kirillov, Kaiming He, Ross Girshick, Carsten Rother, Piotr Dollár, Panoptic segmentation, in: Proceedings of IEEE Conference on Computer Vision and Pattern Recognition, 2019, pp. 9404–9413.

[34] Alexander Kirillov, Evgeny Levinkov, Bjoern Andres, Bogdan Savchynskyy, Carsten Rother, Instancecut: from edges to instances with multicut, in: Proceedings of IEEE Conference on Computer Vision and Pattern Recognition, 2017, pp. 5008–5017.

[35] Sh.u. Kong, Charless C Fowlkes, Recurrent pixel embedding for instance grouping, in: Proceedings of IEEE Conference on Computer Vision and Pattern Recognition, 2018, pp. 9018–9028.

[36] Weicheng Kuo, Anelia Angelova, Jitendra Malik, Tsung-Yi Lin, Shapemask: Learning to segment novel objects by refining shape priors, in: Proceedings of IEEE International Conference on Computer Vision, 2019, pp. 9207–9216.

[37] Fa.had. Lateef, Yassine Ruichek, Survey on semantic segmentation using deep learning techniques, Neurocomputing 338 (2019) 321–348.

[38] Youngwan Lee and Jongyoul Park. Centermask: Real-time anchor-free instance segmentation. Proceedings of IEEE Conference on Computer Vision and Pattern Recognition, pages 13906–13915, 2019.

[39] Qizhu Li, Anurag Arnab, Philip HS Torr, Weakly-and semi-supervised panoptic segmentation, in: Proceedings of European Conference on Computer Vision, 2018, pp. 102–118.

[40] Yi Li, Haozhi Qi, Jifeng Dai, Xiangyang Ji, and Yichen Wei. Fully convolutional instance-aware semantic segmentation, in: Proceedings of IEEE Conference on Computer Vision and Pattern Recognition, 2017, pp. 2359–2367.

[41] Tsung-Yi Lin, Priya Goyal, Ross Girshick, Kaiming He, Piotr Dollár, Focal loss for dense object detection, in: Proceedings of IEEE International Conference on Computer Vision, 2017, pp. 2980–2988.

[42] Tsung-Yi Lin, Michael Maire, Serge Belongie, James Hays, Pietro Perona, Deva Ramanan, Piotr Dollár, and C Lawrence Zitnick. Microsoft coco: Common objects in context. In Proceedings of European Conference on Computer Vision, pages 740–755, 2014.

[43] Sh.u. Liu, Jiaya Jia, Sanja Fidler, Raquel Urtasun, Sgn: Sequential grouping networks for instance segmentation, in: Proceedings of IEEE International Conference on Computer Vision, 2017, pp. 3496–3504.

[44] Jonathan Long, Evan Shelhamer, Trevor Darrell, Fully convolutional networks for semantic segmentation, in: Proceedings of IEEE International Conference on Computer Vision, 2015, pp. 3431–3440.

[45] Davy Neven, Bert De Brabandere, Marc Proesmans, Luc Van Gool, Instance segmentation by jointly optimizing spatial embeddings and clustering bandwidth, in: Proceedings of IEEE Conference on Computer Vision and Pattern Recognition, 2019, pp. 8837–8845.

[46] Alejandro Newell, Zhiao Huang, Jia Deng, Associative embedding: End-to-end learning for joint detection and grouping, in: Proceedings of Advances in Neural Information Processing Systems, 2017, pp. 2277–2287.

[47] David Novotny, Samuel Albanie, Diane Larlus, Andrea Vedaldi, Semi-convolutional operators for instance segmentation, in: Proceedings of European Conference on Computer Vision, 2018, pp. 86–102.

[48] Dim P Papadopoulos, Jasper RR Uijlings, Frank Keller, Vittorio Ferrari, Extreme clicking for efficient object annotation, in: Proceedings of IEEE International Conference on Computer Vision, 2017, pp. 4930–4939.

[49] Shaoqing Ren, Kaiming He, Ross Girshick, Jian Sun, Faster r-cnn: Towards real-time object detection with region proposal networks, in: Proceedings of Advances in Neural Information Processing Systems, 2015, pp. 91–99.

[50] Carsten Rother, Vladimir Kolmogorov, Andrew Blake, rabcutinteractive foreground extraction using iterated graph cuts, ACM Transactions on Graphics 23 (3) (2004) 309–314.

[51] Olga Russakovsky, Jia Deng, Su. Hao, Jonathan Krause, Sanjeev Satheesh, Sean Ma, Zhiheng Huang, Andrej Karpathy, Aditya Khosla, Michael Bernstein, et al.,

Imagenet large scale visual recognition challenge, International Journal of Computer Vision 115 (3) (2015) 211–252.
[52] Chunfeng Song, Yan Huang, Wanli Ouyang, Liang Wang, Box-driven class-wise region masking and filling rate guided loss for weakly supervised semantic segmentation, in: Proceedings of IEEE International Conference on Computer Vision, 2019, pp. 3136–3145.
[53] Towaki Takikawa, David Acuna, Varun Jampani, Sanja Fidler, Gated-scnn: Gated shape cnns for semantic segmentation, in: Proceedings of IEEE International Conference on Computer Vision, 2019, pp. 5229–5238.
[54] Zhi Tian, Chunhua Shen, Hao Chen, Tong He, Fcos: Fully convolutional one-stage object detection, in: Proceedings of IEEE International Conference on Computer Vision, 2019, pp. 9627–9636.
[55] Jun Wang, Shaoguo Wen, Kaixing Chen, Jianghua Yu, Xin Zhou, Peng Gao, Changsheng Li, and Guotong Xie. Semi-supervised active learning for instance segmentation via scoring predictions. In Proceedings of the British Machine Vision Conference, pages 1–13, 2020.
[56] Wu. Fei, Feng Chen, Xiao-Yuan Jing, Hu. Chang-Hui, Qi Ge, Yimu Ji, Dynamic attention network for semantic segmentation, Neurocomputing 384 (2020) 182–191.
[57] Saining Xie, Ross Girshick, Piotr Dollár, Tu. Zhuowen, Kaiming He, Aggregated residual transformations for deep neural networks, in: Proceedings of IEEE Conference on Computer Vision and Pattern Recognition, 2017, pp. 1492–1500.
[58] Hongshan Yu, Zhengeng Yang, Lei Tan, Yaonan Wang, Wei Sun, Mingui Sun, and Yandong Tang. Methods and datasets on semantic segmentation: A review. Neurocomputing, 304:82–103, 2018.
[59] Boxiang Zhang, Wenhui Li, Yuming Hui, Jiayun Liu, Yuanyuan Guan, Mfenet: Multi-level feature enhancement network for real-time semantic segmentation, Neurocomputing 393 (2020) 54–65.
[60] Zhenli Zhang, Xiangyu Zhang, Chao Peng, Xiangyang Xue, and Jian Sun. Exfuse: Enhancing feature fusion for semantic segmentation, in: Proceedings of European Conference on Computer Vision, 2018, pp. 269–284.
[61] Hengshuang Zhao, Jianping Shi, Xiaojuan Qi, Xiaogang Wang, and Jiaya Jia. Pyramid scene parsing network. In Proceedings of IEEE International Conference on Computer Vision, pages 2881–2890, 2017.
[62] Feng Zhou, Hu. Yong, Xukun Shen, Scale-aware spatial pyramid pooling with both encoder-mask and scale-attention for semantic segmentation, Neurocomputing 383 (2020) 174–182.
[63] Yanzhao Zhou, Xin Wang, Jianbin Jiao, Trevor Darrell, and Fisher Yu. Learning saliency propagation for semi-supervised instance segmentation. In Proceedings of IEEE Conference on Computer Vision and Pattern Recognition, pages 10307–10316, 2020.
[64] Yanzhao Zhou, Yi Zhu, Qixiang Ye, Qiang Qiu, and Jianbin Jiao. Weakly supervised instance segmentation using class peak response. In Proceedings of IEEE International Conference on Computer Vision, pages 3791–3800, 2018.
[65] Zhen Zhou, Yan Zhou, Dongli Wang, Mu. Jinzhen, Haibin Zhou, Self-attention feature fusion network for semantic segmentation, Neurocomputing 453 (2021) 50–59.
[66] Yi Zhu, Yanzhao Zhou, Huijuan Xu, Qixiang Ye, David Doermann, and Jianbin Jiao. Learning instance activation maps for weakly supervised instance segmentation. In Proceedings of IEEE Conference on Computer Vision and Pattern Recognition, pages 3116–3125, 2019.